\documentclass{article}

    \PassOptionsToPackage{numbers, compress}{natbib}

\usepackage[preprint]{neurips_2024}

\usepackage[utf8]{inputenc} 
\usepackage[T1]{fontenc}    
\usepackage{hyperref}       
\usepackage{url}            
\usepackage{booktabs}       
\usepackage{amsfonts}       
\usepackage{nicefrac}       
\usepackage{microtype}      
\usepackage[dvipsnames]{xcolor}        
\usepackage{amsmath}
\usepackage{amssymb}
\usepackage{adjustbox}
\usepackage{multirow}
\usepackage{algorithm}
\usepackage{algorithmic}
\usepackage{colortbl}
\usepackage{subcaption}
\usepackage{authblk}

\usepackage[page]{appendix}

\title{Distilling Vision-Language Pretraining for Efficient Cross-Modal Retrieval}

\author[1]{\textbf{Young Kyun Jang}}
\author[2]{\textbf{Donghyun Kim}}
\author[3]{\textbf{Ser-nam Lim}}
\affil[1]{Meta AI}
\affil[2]{Korea University}
\affil[3]{University of Central Florida}

\begin{document}

\maketitle

\begin{abstract}
``Learning to hash'' is a practical solution for efficient retrieval, offering fast search speed and low storage cost. It is widely applied in various applications, such as image-text cross-modal search. In this paper, we explore the potential of enhancing the performance of learning to hash with the proliferation of powerful large pre-trained models, such as Vision-Language Pre-training (VLP) models. We introduce a novel method named Distillation for Cross-Modal Quantization (DCMQ), which leverages the rich semantic knowledge of VLP models to improve hash representation learning. Specifically, we use the VLP as a `teacher' to distill knowledge into a `student' hashing model equipped with codebooks. This process involves the replacement of supervised labels, which are composed of multi-hot vectors and lack semantics, with the rich semantics of VLP. In the end, we apply a transformation termed Normalization with Paired Consistency (NPC) to achieve a discriminative target for distillation. Further, we introduce a new quantization method, Product Quantization with Gumbel (PQG) that promotes balanced codebook learning, thereby improving the retrieval performance. Extensive benchmark testing demonstrates that DCMQ consistently outperforms existing supervised cross-modal hashing approaches, showcasing its significant potential.
\end{abstract}


\section{Introduction}

An efficient search method for large-scale cross-modal retrieval systems is crucial given the exponential rise of multi-modal data, which encompasses diverse formats such as texts, images, audios, and videos. Among various methods, \textit{Hashing} (also known as learning to hash) has emerged as a promising solution. It facilitates Approximate Nearest Neighbor (ANN) search by succinctly encoding high-dimensional data points into compact binary codes \cite{Survey_Hash, PQ, Non_deep_CM, YeUnsupervised, Refining, Non_deep_CM2, QCH, CCQ, CMCQ}. Hashing is a cost-effective solution because it uses binary codes, each only a few bits long, to represent samples. This enables rapid searches through XOR operations or inverted binary computations, quickly determining similarity scores between a query and the gallery.

In recent years, a variety of deep learning to hash methods have been utilized in both single-modal \cite{USPAH,GPQ,SPQ} and multi-modal \cite{DCMH,CMHH,SSAH,CMMQ,DCH-SCR} semantic content searches. The objective of this task is to retrieve samples from the gallery that share the same semantics (such as a classification label) as the query. Here, supervised methods have been shown to yield state of the art performance by utilizing category labels to guide the learning of target hash representations. However, these labels are generally provided as multi-hot vectors that have limited semantic information, suggesting that there is room for performance gains if more detailed semantics were provided.

To address this issue, we turn to Vision-Language Pretraining (VLP) models \cite{Vilbert,CLIP,DeCLIP,SLIP,ALBEF,BLIP}. These models, trained on a vast number of image-text pairs from diverse domains and datasets, encapsulate rich semantic similarities between images and texts. However, the significant computational demands of VLP models for both training and inference hinder their direct application in building efficient retrieval systems. Furthermore, fine-tuning VLP models for specific retrieval tasks risks impairing their fine-grained semantic understanding.

\begin{figure}[!t]
\centering
  \subcaptionbox{Supervised Deep Cross-Modal Hashing
  \label{fig:intro_a}}{\vspace{1.5em}\includegraphics[width=0.43\linewidth]{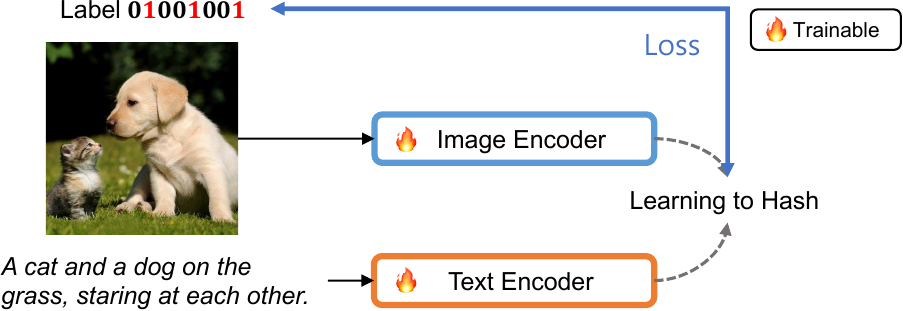}}
  \subcaptionbox{Distillation for Cross-Modal Quantization (Ours)
  \label{fig:intro_b}}{\includegraphics[width=0.56\linewidth]{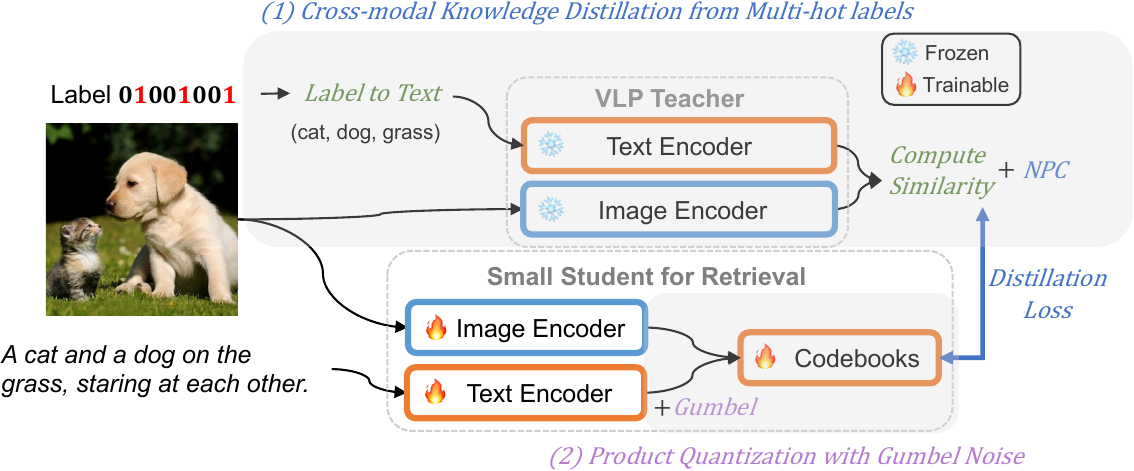}}
\caption{A comparison of the training process between (a) existing supervised deep hashing and (b) our DCMQ. Supervised labels, which only indicate category presence, lack detailed cross-modal semantic information. To address this, we convert these labels into text and employ Vision-Language Pretraining (VLP) to compute cross-modal similarity between images and labels, thereby forming the target distribution for knowledge distillation. Performance gains from this process are achieved through the introduction of two crucial techniques: Normalization with Paired Consistency (NPC) and Product Quantization with Gumbel noise (PQG).
}
\label{fig:intro}
\end{figure}

To overcome these challenges, we introduce a unique method, \textit{\textbf{D}istillation for \textbf{C}ross-\textbf{M}odal \textbf{Q}uantization} (DCMQ). DCMQ distills the knowledge of VLP into smaller encoders, improving both retrieval accuracy and computational efficiency. In DCMQ, the `teacher' VLP guides `student' model, which is designed for hashing, as shown in Figure \ref{fig:intro_b}. A straightforward approach to distilling VLP knowledge for cross-modal hash learning would involve forwarding an image and its corresponding text to the VLP, using its output as a supervisory signal. However, text can often be noisy, while label names tend to more reliably capture object semantics. Conversely, object label names may fail to capture semantics such as `walking' or `talking', which are often present in text and are equally important for the student model to learn. Therefore, we propose a training recipe designed to harness the strengths of both text and labels.

\begin{figure}[!t]
\centering
\includegraphics[width=0.7\linewidth]{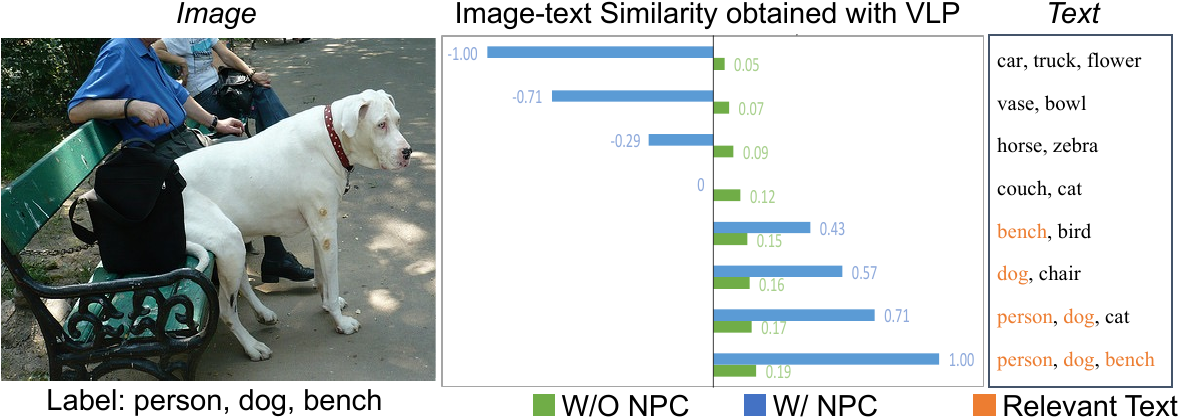}
\caption{An example of how NPC impacts the cosine similarity scores between image and text output embeddings of VLP. \textcolor{OliveGreen}{Without NPC}, the scores are within the range of 0.05 to 0.19, which do not effectively distinguish between relevant (positive) and non-relevant (negative) text. However, \textcolor{RoyalBlue}{with NPC}, the scores are normalized within the range of -1.0 to 1.0, showing enhanced semantic relevance between image and texts.}
\label{fig:NPQ_example}
\end{figure}

Specifically, the training process begins by converting a multi-hot label associated with an image into its category name, thereby creating the text data that effectively captures object semantics. The VLP encoders then process this paired image-text data to yield corresponding embeddings. From these, we derive a cross-modal target similarity matrix, $\mathbf{T}$, that we will use to teach the student to hash. However, we observe that directly using $\mathbf{T}$ performs worse than utilizing original labels because most of the similarity scores are densely distributed within a short range, showing less discrimination as depicted in Figure \ref{fig:NPQ_example}. Therefore, we introduce a transformation termed \textit{Normalization with Paired Consistency} (NPC) that broadly redistributes scores to produce a well-separated $\mathbf{T}$ distribution, which we show in this paper is crucial for learning discriminative hash representation. Observing the green bars (without NPC) and blue bars (with NPC) in Figure \ref{fig:NPQ_example}, it is clear that the blue bars yield more distinct similarities for relevant text as opposed to negative text.

For the student hashing model, we adopt a Product Quantization (PQ) \cite{PQ}-based deep learning approach. This approach is popular due to its expressive power, which stems from using a multi-codebook to represent gallery samples in binary codes. However, deep learning for PQ methods \cite{PQN,GPQ,SPQ} often leads to overfitting to certain codewords, resulting in sub-optimal codeword usage and limiting generalization. To address this issue, we introduce a novel technique, \textit{PQ with Gumbel} (PQG). PQG employs Gumbel noise \cite{Gumbel} to regularize codeword selection during training, promoting balanced codeword contribution and enhancing the retrieval model's generalization capacity. This increased capacity enables the model to capture additional semantics present in the caption, as the student model uses the text as input.

In order to validate DCMQ, we conducted extensive experiments on various image-text cross-modal retrieval benchmark datasets. We utilize a diverse set of VLP models, which differ in parameter scale, architectural design, and learning strategies, to demonstrate DCMQ's broad applicability and robust effectiveness. The results confirm DCMQ's efficiency and its outstanding retrieval performance in line with the state-of-the-art across cross-modal protocols.

\section{Related Work}
\label{sec:related work}

\noindent \textbf{Deep Learning to Hash.} Learn to hash has taken advantage of deep learning to provide training signals with designed objectives from multi-media data such as text, image, audio and video \cite{Text_hashing, Refining, YeUnsupervised,DSDH,CSQ,DHD,Audio_hashing,Video1,Video2,Video3}. Deep supervised hashing methods have been extensively studied in various fields and remains an active field of research today. For example, tasks like face retrieval \cite{DCBH} and biometric authentication \cite{DFC}, or compression \cite{DiffPQ_compression} have all taken advantage of deep hash code representations. Product Quantization \cite{PQ} (PQ) based deep approaches produce real-valued hash representations and is one of the mainstream learn to hash methods. Specifically, researchers have verified the effectiveness of supervised \cite{PQN,EtePQ,BeyPQ}, semi-supervised \cite{GPQ} and unsupervised \cite{SPQ, YeUnsupervised} deep PQ methods with outstanding performances in content-based retrieval.

\noindent \textbf{Learning to Hash for Cross-Modal Retrieval.} For the purpose of generating hash codes that are aligned between data of different modalities, both non-deep \cite{Non_deep_CM,Non_deep_CM2,QCH,CCQ,CMCQ} and deep learning \cite{MM_neuro, Video_text, Semantic_driven} methods have been explored. In particular, for content-based cross-modal search between images and texts, deep supervised \cite{DCMH,RDCMH,CMHH,Triplet_DCM,CMMQ,DCH-SCR} methods show promising performance by utilizing additional multi-hot annotation vectors. Deep unsupervised methods \cite{USPAH,ASSPH,UGACMH,SSAH,Creating,Joint,Graph-neighbor} have also been considered which take training signals from data reconstruction or constructed graphs to learn mutual alignment between modalities. Despite these advances, this paper shows that we can capitalize on the knowledge of VLP to achieve better hash representations.

\noindent \textbf{Distilling Vision-Language Pretraining.} Connecting vast amounts of images and texts with deep vision and language encoders have shown great success in building high-quality VLP models \cite{Vilbert,CLIP,SLIP,DeCLIP}. In order to take advantage of its intelligence while maintaining model efficiency, knowledge distillation strategy \cite{KD_survey} has been introduced to build smaller VLP models \cite{Compressing_VLP} or to facilitate training with object information \cite{KD-VLP}. Some methods including \cite{Alignment_codebook} use self-distillation which adopts a self-supervised approach to learn a representation codebook by identifying stable alignments between modalities using prototypes, and \cite{Robust}, which modifies the similarity matrix used as the training target to address the challenge of noisy vision-language alignments. To the best of our knowledge, our work is the first to teach a significantly smaller student model with the knowledge of a VLP while also acquiring the ability to hash. Unlike conventional distillation methods that aim to train continuous student models, hashing involves an indifferentiable quantization process, making the distillation in our case non-trivial.

\section{Method}
\label{sec:method}

\subsection{Overview}
\label{sec:method:overview}

The goal of cross-modal retrieval is mapping images and text (captions) into a shared embedding space, and finding alignment between the same semantic contents. Consider a vision-language dataset with $N_D$ samples given as $\mathcal{D}=\{\mathcal{I}_{n},\mathcal{T}_{n},y_{n}\}^{N_D}_{n=1}$, where $\mathcal{I}_{n}$, $\mathcal{T}_{n}$, and $y_{n}$ denotes $n$-th image, text, and the corresponding multi-hot encoded label, respectively. With image encoder $\mathcal{E}^{i}$ and text encoder $\mathcal{E}^{t}$, two separate image and text feature vector embedding $\mathbf{x}^i$ and $\mathbf{x}^t$ of the same $\mathrm{D}$-dimensionality are produced as outputs.

We configure $M$-codebooks of $K$-codewords for deep PQ parameter notate as $\mathcal{C}=\{\mathbf{C}_m\}^{M}_{m=1}$ where $\mathbf{C}_m=\{\mathbf{\hat{c}}_{mk}\}^{K}_{k=1}$ of $\mathbf{\hat{c}}_{mk}\subset R^{\mathrm{D}/M}$, to collect discriminative cross-modal representation. The entire trainable components of DCMQ, $\mathcal{E}^{i}$, $\mathcal{E}^{t}$ and $\mathcal{C}$, are jointly trained in a single framework, end-to-end manner. In order to update model (student) parameters with supervision, cosine similarity between $N$ samples of paired $\mathbf{x}^i$-$\mathbf{x}^t$ and their corresponding soft-quantized embeddings derived from the codebooks (Section~\ref{sec:method:PQG}) are computed across modalities (Section \ref{sec:method:Finding Cross-Modal Alignment}). These similarity score outputs from student are then supervised by \textit{pre-computed} target similarity matrix $\mathbf{T}$ from VLP model (teacher, Section \ref{sec:method:VLP as supervised teacher}).

\subsection{VLP as Supervisory Teacher with NPC}
\label{sec:method:VLP as supervised teacher}

\begin{figure}[!t]
\centering
\begin{minipage}{.55\textwidth}
  \centering
  \subcaptionbox{Before NPC
  \label{fig:NPC_before}}{\includegraphics[height=3.5cm]{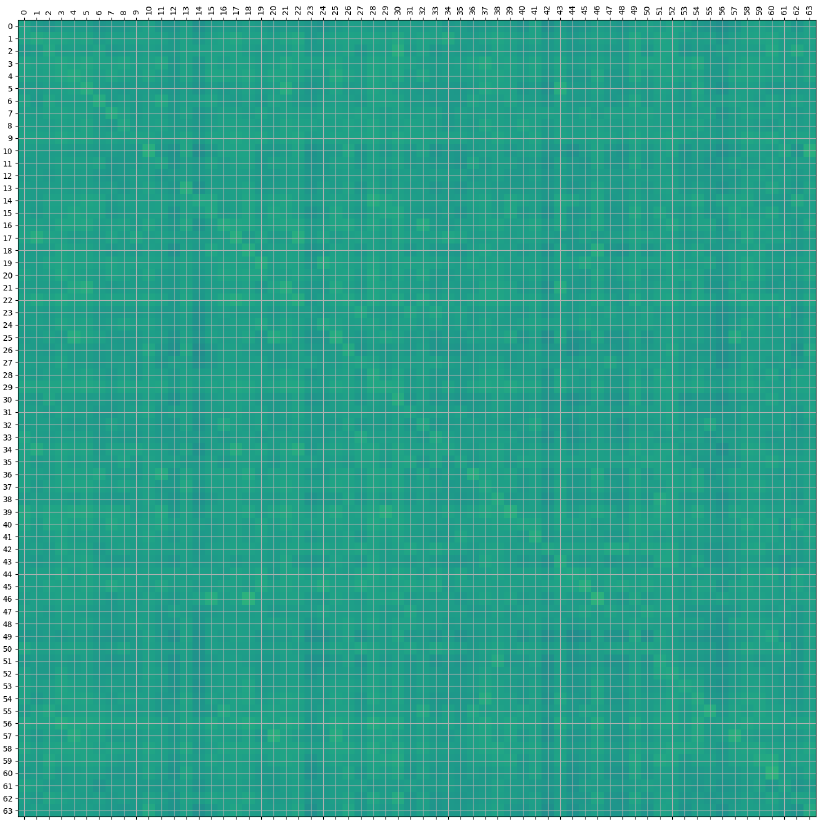}}
  \subcaptionbox{After NPC
  \label{fig:NPC_after}}{\includegraphics[height=3.5cm]{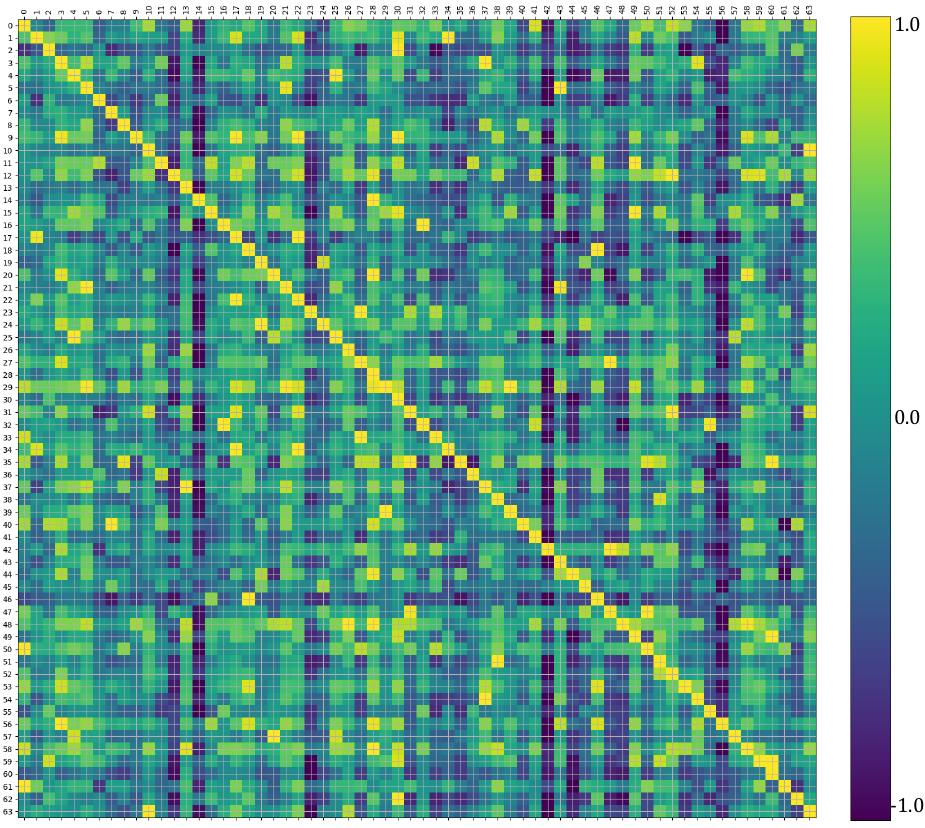}}
\caption{An example similarity matrix computed with 64 VLP image and text embeddings. After NPC, scores become more dominant while keeping paired consistency.}
\label{fig:NPC}
\end{minipage}\hspace{0.5em}%
\begin{minipage}{.42\textwidth}
  \centering
  \begin{algorithm}[H]
  \caption{NPC algorithm.}
  \begin{algorithmic}[1]
  \label{algorithm:NPC}
  \small

  \REQUIRE $\mathbf{T}=\left[\mathbf{s}_1, ..., \mathbf{s}_N\right], \text{ where } \mathbf{s}_i =\left[s_{i1}, ..., s_{iN}\right]$ and $\mathbf{s}_{i}$ is $i$-th row of $\mathbf{T}$, and ${s}_{ij}$ is $j$-th element of $\mathbf{s}_{i}$
  \FOR{$i$ in \{1, ..., $N$\}}
  \STATE ${s}_{\mathrm{M}}=\max{(\mathbf{s}_i)}, {s}_{\mathrm{m}}=\min{(\mathbf{s}_i)}$
  \STATE ${a}_{i}=\frac{2.0}{{s}_{\mathrm{M}}-{s}_{\mathrm{m}}},{b}_{i}=-\frac{{s}_{\mathrm{M}}+{s}_{\mathrm{m}}}{{s}_{\mathrm{M}}-{s}_{\mathrm{m}}}$
  \FOR{$j$ in \{1, ..., $N$\}}
  \STATE ${s}_{ij}={a}_{i}\times {s}_{ij} + {b}_{i}$
  \IF{$i=j$}
  \STATE ${s}_{ij}=1.0$
  \ENDIF
  \ENDFOR
  \ENDFOR
  \ENSURE Updated $\mathbf{T}$

  \end{algorithmic}
  \end{algorithm}
\end{minipage}
\end{figure}

To harness the power of the semantic knowledge captured in a VLP model $\{\mathcal{V}^{i},\mathcal{V}^{t}\}$, where $\mathcal{V}^{i}$ and $\mathcal{V}^{t}$ denotes the image and text encoder respectively, $\mathcal{I}$ and $y$ are forwarded to produce embeddings as $\mathbf{v}^{i}=\mathcal{V}^{i}(\mathcal{I})$ and $\mathbf{v}^{t}=\mathcal{V}^{t}(\hat{y})$. Note that $\hat{y}$ is the text-translated version of $y$ where the labels are replaced with the corresponding category names, and all the embeddings are $l2$-normalized. 
We then calculate a similarity matrix based on the cross-modal similarities between $N$ pairs of image and text with a collection of VLP embeddings $\mathbf{V}^{i}=[\mathbf{v}^{i}_1 , ... , \mathbf{v}^{i}_N]$ and $\mathbf{V}^{t}=[\mathbf{v}^{t}_1 , ... , \mathbf{v}^{t}_N]$ as:

\begin{align}
\mathbf{T} = \mathbf{V}^{i} \cdot \left(\mathbf{V}^{t}\right)^\intercal
\label{equation:compute similarity matrix}
\end{align}

\noindent where $\mathbf{T}$ is a similarity matrix that stands for cosine similarity scores among embeddings. 
However, initial observations revealed that the cosine similarity scores in $\mathbf{T}$ are distributed only within a small range as illustrated in Figure \ref{fig:NPC_before}, likely because VLP is trained on far more samples than the retrieval training set. Therefore, adapting $\mathbf{T}$ to the target domain is essential, so we introduce a Normalization with Paired Consistency (NPC) technique given in Algorithm \ref{algorithm:NPC}.

\noindent Note that, NPC maps every element of $\mathbf{s}_{i}$ into a linear function of ${a}_{i}$ slope and ${b}_{i}$ y-axis intercept, and re-scaling them to -1.0 to 1.0. In order to maintain consistency between image and text pairs, the diagonal components of $\mathbf{T}$ are set to have the maximum (1.0) value, avoiding miss-alignment between modalities. Figure \ref{fig:NPC_after} shows the positive effect of this normalization. $\mathbf{T}$ is then used to supervise the training of the components ($\mathcal{E}^{i}$, $\mathcal{E}^{t}$ and $\mathcal{C}$).

Such a VLP supervision scheme has several advantages. First, the entire VLP embeddings can be obtained and saved offline once, so we only need a single inference of large VLP model. Second, a simple matrix multiplication with normalization is the only additional computational load, which is efficient. Lastly, VLP supervision has the effect of transferring the rich semantic understandings, which is better than multi-hot annotations.

\subsection{Product Quantization with Gumbel}
\label{sec:method:PQG}

We adopt PQ \cite{PQ} to find codewords (centroids) for quantization and learn hash representation. Inspired by previous deep PQ works \cite{PQN,GPQ,SPQ} that approximates quantization (argmax) with differentiable Softmax operation during training, we propose a Product Quantization with Gumbel (PQG) scheme which is derived to boost the influence of non-attentive codewords as:

\begin{align}
\mathbf{\hat{z}}_m= \mathcal{A}\left(\mathbf{\hat{x}}_m,\mathbf{C}_m\right) + \lambda\cdot \mathcal{A}_g\left(\mathbf{\hat{x}}_m,\mathbf{C}_m\right),
\label{equation:PQG}
\end{align}

\noindent where $\lambda$ is balancing hyper-parameter and $\mathbf{\hat{z}}_m$ is a soft quantized output of $\mathbf{\hat{x}}_m$, which is a $m$-th sub-vector obtained by slicing $\mathbf{x}$ into $M$ pieces as $\left[\mathbf{\hat{x}}_1, ..., \mathbf{\hat{x}}_M\right]$. $\mathcal{A}$ denotes attention module, taking query as $\mathbf{\hat{x}}_m$ and key, value as $\mathbf{\hat{c}}_{mk}$, defined as:

\begin{align}
\mathcal{A}=\sum_{k}^{K}\left(\mathbb{S}\left(\mathcal{S}_c\left(\mathbf{\hat{x}}_m,\mathbf{\hat{c}}_{mk}\right)\right)\cdot\mathbf{\hat{c}}_{mk}\right)
\label{equation:Attention}
\end{align}

\noindent where $\mathcal{S}_c(\cdot,\cdot)$ implies cosine similarity between inputs. The argument $\mathbb{S}(\cdot)$ denotes normalized-temperature Softmax and can be replaced to introduce randomness in $\mathcal{A}$, as $\mathcal{A}_g$, with Gumbel-Softmax \cite{Gumbel} $\mathbb{S}_g(\cdot)$, which are formulated as:

\begin{align}
\mathbb{S}(\mathcal{S}_c;\tau_{\mathbb{S}})=\frac{e^{\mathcal{S}_c\left(\mathbf{\hat{x}}_m,\mathbf{\hat{c}}_{mk}\right)/\tau_{\mathbb{S}}}}{\sum_{k'}^{K}e^{\mathcal{S}_c\left(\mathbf{\hat{x}}_m,\mathbf{\hat{c}}_{mk'}\right)/\tau_{\mathbb{S}}}}
\label{equation:Softmax}
\end{align}

\begin{align}
\mathbb{S}_g(\mathcal{S}_c;\tau_{\mathbb{S}_g})=\frac{e^{\left(\mathcal{S}_c\left(\mathbf{\hat{x}}_m,\mathbf{\hat{c}}_{mk}\right)+g_k\right)/\tau_{\mathbb{S}_g}}}{\sum_{k'}^{K}e^{\left(\mathcal{S}_c\left(\mathbf{\hat{x}}_m,\mathbf{\hat{c}}_{mk'}\right)+g_{k'}\right)/\tau_{\mathbb{S}_g}}}
\label{equation:GumbelSoftmax}
\end{align}

\noindent where $\tau_\mathbb{S}$ and $\tau_{\mathbb{S}_g}$ denote temperature hyper-parameters and $g_k$ denotes noise sampled from standard Gumbel distribution ($\mu=0$ and $\beta=1$). Here, we integrate deterministic $\mathbb{S}(\cdot)$ and stochastic $\mathbb{S}_g(\cdot)$ into the quantization. Doing so helps to increase the contribution of every codeword for better generalization instead of merely focusing on the most attentive codewords.

Finally, Equation \ref{equation:PQG} is applied on every $\mathbf{\hat{x}}_m$ of image and text embeddings to generate $\mathbf{\hat{z}}_m$, and then, all $\mathbf{\hat{z}}_m$ are concatenated to produce a soft-quantized embedding $\mathbf{z}=\left[\mathbf{\hat{z}}_1, ..., \mathbf{\hat{z}}_M\right]$ as shown in Figure \ref{fig:PQG}. Specifically, even though image and text is paired and sharing the same codewords of the codebook, Gumbel-trick provides extra room for representing $\mathbf{z}^i$ and $\mathbf{z}^t$, increasing the robustness in the alignment of modalities. Moreover, the cross-modal representation, integrated within a continuous coding space of $\mathcal{C}$, facilitates improved hash representation learning for more accurate retrieval. Section \ref{subsec:Ablation Study} presents our ablation study on the PQG.

\subsection{Finding Cross-Modal Alignment}
\label{sec:method:Finding Cross-Modal Alignment}

\begin{figure}[!t]
\centering
\subcaptionbox{Product Quantization with Gumbel (PQG).
}{\includegraphics[width=0.55\linewidth]{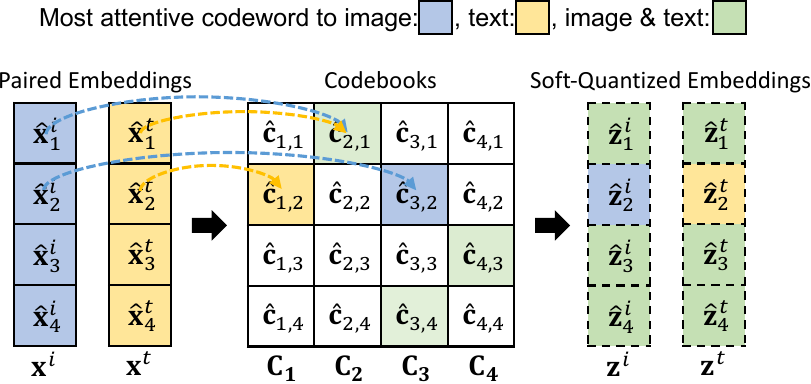}}
\subcaptionbox{Cross-modal joint training.
\label{fig:Training}}{\vspace{2em}\includegraphics[width=0.43\linewidth]{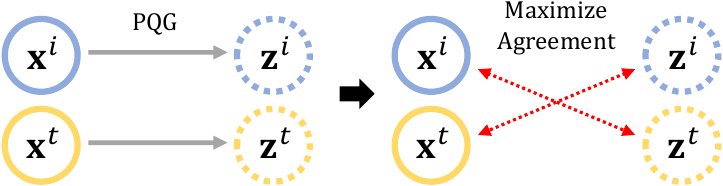}}

\caption{(a) PQG on multi-modal data. $l$-th sub-vector of Soft-quantized embeddings; $\mathbf{\hat{z}}^i_{l}$ and $\mathbf{\hat{z}}^t_{l}$ are created by a combination of codewords in $l$-th codebook $\mathbf{C}_l$. (b) Joint training of image ($\mathbf{x}^i$, $\mathbf{z}^i$) and text ($\mathbf{x}^t$, $\mathbf{z}^t$) embeddings in order to obtain cross-modal alignment between the original and quantized representation. }
\label{fig:PQG}
\end{figure}

To train all the learnable components, namely image and text encoders ($\mathcal{E}^i$, $\mathcal{E}^t$) and $\mathcal{C}$ with VLP supervision, we design two training objectives as:

\begin{align}
\mathcal{L}_{i2t} &= \mathcal{L}_{ce}\left(\mathbb{S}\left(\mathbf{C}^{i2t};\tau_{ce}\right),  \mathbb{S}\left(\mathbf{T};\tau_{ce}\right)\right) \\
\mathcal{L}_{t2i} &= \mathcal{L}_{ce}\left(\mathbb{S}\left(\mathbf{C}^{t2i};\tau_{ce}\right),  \mathbb{S}\left(\left(\mathbf{T}\right)^\intercal;\tau_{ce}\right)\right)
\label{equation:cross_modal_training}
\end{align}

\noindent where $\mathcal{L}_{ce}$ is a standard cross entropy loss function for training batch, and $\mathbb{S}(\cdot)$ is Softmax, same as Equation \ref{equation:Softmax} but with different temperature $\tau_{ce}$. A collection of embeddings are utilized to get image-to-text ($i2t$) similarity matrix $\mathbf{C}^{i2t}=\mathbf{Z}^i\cdot \left(\mathbf{X}^t\right)^\intercal$, where $\mathbf{Z}^{t}=[\mathbf{z}^{t}_1 , ... , \mathbf{z}^{t}_N]$, $\mathbf{X}^{i}=[\mathbf{x}^{i}_1 , ... , \mathbf{x}^{i}_N]$, and vice versa for text-to-image ($t2i$).

As shown in Figure \ref{fig:Training}, we prioritize maximizing the similarity between $\mathbf{Z}^{i}$ and $\mathbf{X}^{t}$, as well as between $\mathbf{Z}^{t}$ and $\mathbf{X}^{i}$, over the similarity between $\mathbf{Z}^{i}$ and $\mathbf{Z}^{t}$. Empirical observations suggest this approach yields superior results, which is in line with findings reported in \cite{SPQ}, and we report results in Appendix \ref{sec:appendix:Effect of joint training}.


\subsection{Inference}
\label{sec:Inference}

In order to perform image to text retrieval (similar procedure applies for text to image), the text gallery is built as follows. First, a given text, $\mathcal{T}_g$, is forwarded to $\mathcal{E}^t$ to produce $\mathbf{x}^t_g$, which is then sliced to generate $M$ sub-vectors. Second, we find the indices of the nearest codewords from the codebooks corresponding to each sub-vector. Lastly, we convert the indices into binary codes and concatenate them to generate a binary code. This process (inverted binary indexing) is repeated for all the text to be stored in the gallery. During retrieval, a query image $\mathcal{I}_q$ is forwarded to $\mathcal{E}^i$ and an embedding $\mathbf{x}^i_q$ is generated. $\mathbf{x}^i_q$ is sliced into $M$ sub-vectors and their cosine similarities with every codewords in the codebooks are calculated to create a look-up table. The asymmetric distance between query and gallery is computed and accelerated by summing up the look-up results.


\section{Experiments}
\label{sec:Experiments}

\subsection{Settings}
\label{subsec:Settings}

We ensure all artifacts used in our paper adhere to their specific licensing terms, permitting research use. Additionally, we confirm the data does not contain personal identifiers or offensive content.

\paragraph{Datasets.} In order to make a fair comparison with previous works \cite{DCMH,CMMQ}, we evaluate our method on three datasets; MS COCO, MIRFlickr, and NUS-WIDE used by these earlier works. Detailed configuration are found in Appendix \ref{sec:appendix:dataset}.

\noindent \textbf{Evaluation Metrics.} We follow the same practice in community \cite{CMMQ} for evaluation. In terms of cross-modal retrieval (Text query to Image gallery $T\rightarrow{I}$, and Image query to Text gallery $I\rightarrow{T}$) evaluation, we employ metrics as: mean Average Precision (mAP) to $N_{mAP}$-ranked results, Top $N_{Top}$-precision curve to at top $N_{Top}$, and Recall at $N_{Recall}$. We set $\{N_{mAP}, N_{Top}, N_{Recall}\}=\{5,000, 1,000, 1000\}$, and set the target number of bits $N_{bits}$ as 32, 64 and 128 for evaluation.

\noindent \textbf{Implementation details.} For a fair comparison with existing methods, we fix the backbone of image and text encoder of the student as ImageNet pretrained ResNet18 (RN18) \cite{ResNet} and three layer Multi-Layer Perceptron (MLP) respectively for DCMQ. Note that, just like earlier work, we use a MLP for the text encoder because the text for the datasets are provided as BoW feature vectors instead of the actual text. We vary the number of codebooks $M=\{8,16,32\}$ to match the target number of bits as $N_{bits}=M \times \log_2 K$, while fixing the number of codewords $K$ at $2^4$. We balance the contribution between $\mathbb{S}$ and $\mathbb{S}_g$ in PQG equally by setting $\lambda$ as 1.0 for the noisy (due to missing labels) MIRFlickr and NUS-WIDE, but adjust $\lambda$ to 0.5 for the less noisy MS COCO. Temperature hyper-parameters are set accordingly as $\{\tau_{\mathbb{S}},\tau_{\mathbb{S_g}},\tau_{ce}\} = \{0.2, 1.0, 0.2\}$. CLIP \cite{CLIP}, DeCLIP \cite{DeCLIP} and their variants are used as the teachers, where ViT-B32 is chosen as the default image encoder ($\text{DCMQ}^{\textrm{CLIP-ViT-B32}}$, $\text{DCMQ}^{\textrm{DeCLIP-ViT-B32}}$), and all VLP embeddings are prepared offline once. We adopt Adam optimizer \cite{Adam} with an initial learning rate $\gamma$ of $1\times10^{-5}$, and reduce $\gamma$ by a tenth after 10 out of 20 epochs on a NVIDIA A100-40G GPU.

For the purpose of retrieval, final fully-connected layer of encoders are replaced with a new layer that outputs $\mathrm{D}=256$ and $N_{bits}$ dimensionality embeddings for PQ and hashing methods respectively. Especially for the non-deep PQ \cite{PQ} and OPQ \cite{OPQ}, we forward images to the same RN18 encoder without new layer, and apply Principal Component Analysis (PCA) to obtain output dimension of $\mathrm{D}$. Additionally for text BoWs, we directly apply another PCA to generate the same $\mathrm{D}$-dimensional outputs for codebook training. For deep learning based learn to hash methods, we carefully re-implement based on the codes provided by the authors of DCMH\cite{DCMH} and SSAH\cite{SSAH}, and we carefully implement CMHH \cite{CMHH} and CMMQ \cite{CMMQ} by following the training details provided in the original papers. The other PQ methods in our work also follow the same hyper-parameter setup with DCMQ. Upon replacing multi-hot label annotations with actual category names, we empirically discovered that simply using spaces between names yielded the best results. Consequently, we opted not to use special prompts in this context. We utilize Intel(R) Xeon(R) Platinum 8275CL CPU @3.00GHz in all experiments.

\subsection{Comparing with Existing Methods}
\label{subsec:Comparison Results with Exisiting Methods}

\begin{table*}[!t]
\centering
\caption{mAP(@5000) scores of all methods for different bits on three cross-modal datasets.}
\begin{adjustbox}{width=0.99\textwidth}
\begin{tabular}{c|l|c|c|c|c|c|c|c|c|c}
\toprule

\multirow{2}{*}{Task}    &  \multicolumn{1}{c|}{\multirow{2}{*}{Method}} & \multicolumn{3}{c|}{MS COCO} & \multicolumn{3}{c|}{MIRFlickr} & \multicolumn{3}{c}{NUS-WIDE} \\ \cmidrule{3-11} 
                           &    \multicolumn{1}{c|}{}                        & 32-bit   & 64-bit  & 128-bit  & 32-bit   & 64-bit  & 128-bit  & 32-bit  & 64-bit  & 128-bit  \\ \midrule

\multirow{8}{*}{$T\rightarrow{I}$}   & PQ \cite{PQ}   & 0.308  & 0.341   & 0.370 & 0.573   & 	0.565 & 	0.554  & 0.304  & 0.315  & 	0.358   \\
                       &            OPQ \cite{OPQ}        & 	0.504   & 	0.406   & 0.396          & 0.637  & 	0.643  & 	0.642  & 	0.324   & 	0.313 & 	0.321  \\
                       &                  DCMH \cite{DCMH}    & 	0.523  & 0.533   & 0.535       &  0.702    & 	0.715  & 0.719  & 0.582   & 	0.592   & 0.587    \\
  &                  SSAH   \cite{SSAH}     & 	0.541  & 0.542   & 0.542     & 0.719    & 	0.720  & 0.722  & 0.592   & 	0.606   & 0.598     \\ 
  &                  CMHH \cite{CMHH}    & 	0.536  & 0.543   & 0.539       & 0.720    & 	0.722  & 0.715  & 0.589   & 	0.588   & 0.595     \\ 
&                  CMMQ \cite{CMMQ}    & 	0.548  & 0.565   & 0.559      & 0.733    & 	0.739  & 0.742  & 0.611   & 	0.615   & 0.620      \\ 
&                  DCH-SCR \cite{DCH-SCR}    & 	0.683  & 0.693   & 0.690   & 0.796   & 	0.802   & 0.804   & 0.723    & 	0.734  & 0.745        \\ 
  &    \cellcolor{lime!15}  $\text{DCMQ}^{\textrm{CLIP-ViT-B32}}$  & \cellcolor{lime!15} 	\textbf{0.849}	 & \cellcolor{lime!15} \textbf{0.859}    & \cellcolor{lime!15} \textbf{0.862}   & \cellcolor{lime!15} 0.812   & \cellcolor{lime!15}0.815  & \cellcolor{lime!15} 0.812  & \cellcolor{lime!15} 0.748    & \cellcolor{lime!15} 0.761  &\cellcolor{lime!15}  0.760   \\
  &    \cellcolor{lime!15}  $\text{DCMQ}^{\textrm{DeCLIP-ViT-B32}}$ & \cellcolor{lime!15} 	0.843	 & \cellcolor{lime!15} 0.856   & \cellcolor{lime!15} 0.861   & \cellcolor{lime!15} \textbf{0.817}	  & \cellcolor{lime!15} \textbf{0.818}  & \cellcolor{lime!15} 	\textbf{0.822}  & \cellcolor{lime!15} \textbf{0.750}   & \cellcolor{lime!15} \textbf{0.761}  &\cellcolor{lime!15}  	\textbf{0.771}   \\ 

\midrule

\multirow{8}{*}{$I\rightarrow{T}$}   & PQ \cite{PQ}   & 	0.370	   & 0.369	  & 0.373   & 0.550	    & 0.550  & 	0.553  & 0.314	    & 0.309	   & 0.308   \\
                       &            OPQ \cite{OPQ}      & 0.370  & 	0.370   & 0.333          & 0.553    & 0.552   & 	0.538 & 0.315   & 	0.296  & 0.316     \\
                       &                  DCMH \cite{DCMH}    & 	0.602  & 0.611  & 0.612      & 0.724    & 0.740  & 0.741   & 0.589    & 	0.621    & 	0.618   \\
  &                  SSAH \cite{SSAH}    & 	0.633  & 0.634  & 0.638      & 0.741    & 0.755  & 	0.750   & 0.602    & 	0.629    & 	0.644      \\ 
  &                  CMHH \cite{CMHH}      & 	0.629  & 0.627  & 0.636      & 0.733    & 0.750  & 	0.753   & 0.599    & 	0.635    & 	0.640    \\ 
&                  CMMQ \cite{CMMQ}    & 	0.633  & 0.641  & 0.648       & 0.742    & 0.756  & 	0.765   & 0.632    & 	0.641    & 	0.655      \\ 
&                  DCH-SCR \cite{DCH-SCR}    & 	0.685  & 0.690   & 0.689   & 0.823   & 	0.833   & 0.835   & 0.730    & 	0.731  & 0.728        \\ 
  &    \cellcolor{lime!15}  $\text{DCMQ}^{\textrm{CLIP-ViT-B32}}$ & \cellcolor{lime!15} \textbf{0.834}	 & \cellcolor{lime!15} \textbf{0.837}   & \cellcolor{lime!15} \textbf{0.841}  & \cellcolor{lime!15} 0.852   & \cellcolor{lime!15} 0.851  & \cellcolor{lime!15} 0.850  & \cellcolor{lime!15} \textbf{0.754}   & \cellcolor{lime!15} \textbf{0.764}  &\cellcolor{lime!15}  \textbf{0.768}    \\
  &    \cellcolor{lime!15}  $\text{DCMQ}^{\textrm{DeCLIP-ViT-B32}}$ & \cellcolor{lime!15} 	0.819	 & \cellcolor{lime!15} 	0.826   & \cellcolor{lime!15} 	0.821   & \cellcolor{lime!15} \textbf{0.854}	   & \cellcolor{lime!15} \textbf{0.856}  & \cellcolor{lime!15} \textbf{0.852}  & \cellcolor{lime!15} 0.745    & \cellcolor{lime!15} 0.747  &\cellcolor{lime!15}  0.742   \\

\bottomrule
\end{tabular}
\end{adjustbox}
\label{table:Comparisons}
\end{table*}

\begin{figure*}[!t]
\centering
  \subcaptionbox{$T\rightarrow{I}$
  \label{Top1000T2I}}{\includegraphics[height=3.2cm]{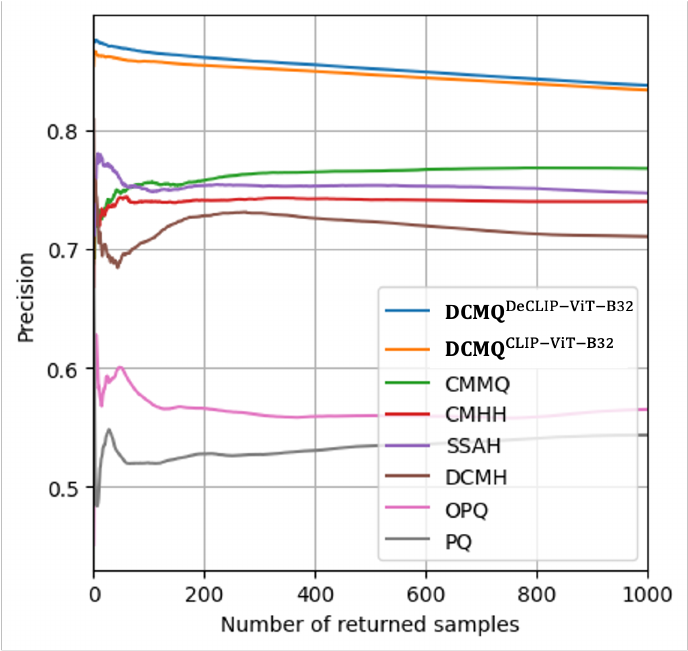}}
  \subcaptionbox{$I\rightarrow{T}$
  \label{Top1000I2T}}{\includegraphics[height=3.2cm]{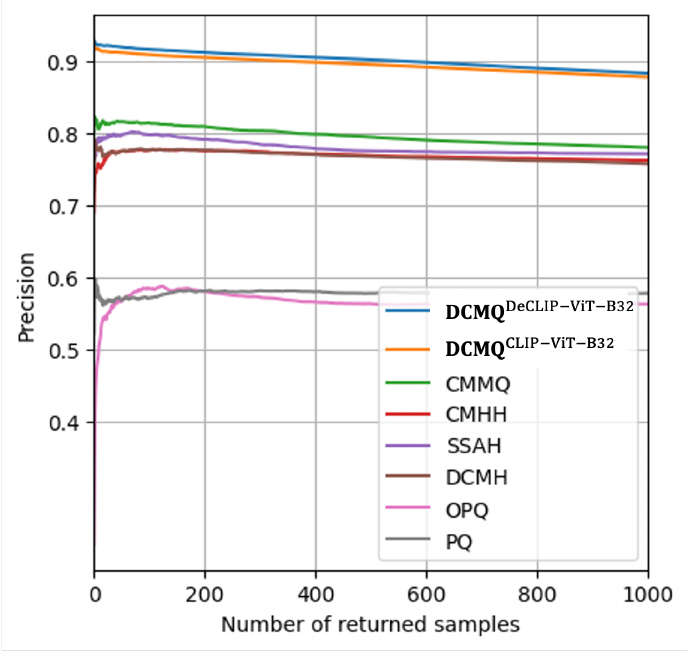}}
  \subcaptionbox{$T\rightarrow{I}$
  \label{Recall100T2I}}{\includegraphics[height=3.2cm]{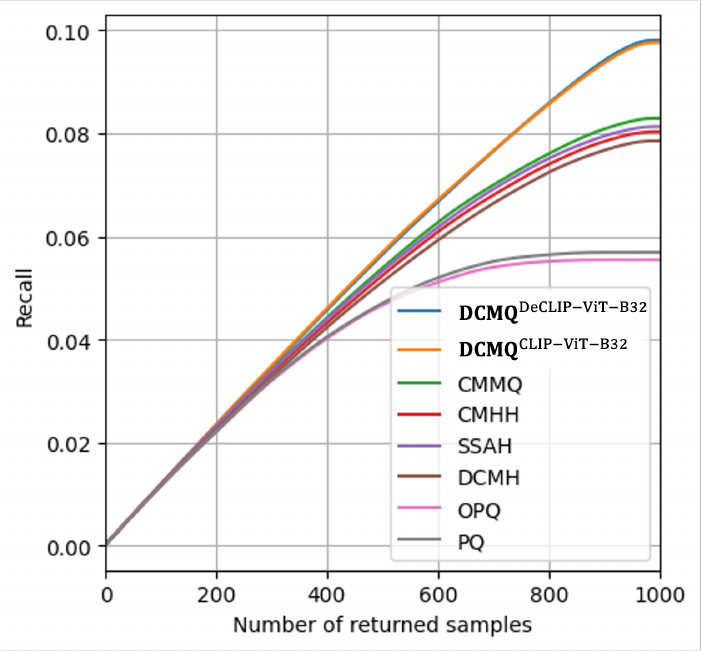}}
  \subcaptionbox{$I\rightarrow{T}$
  \label{Recall100I2T}}{\includegraphics[height=3.2cm]{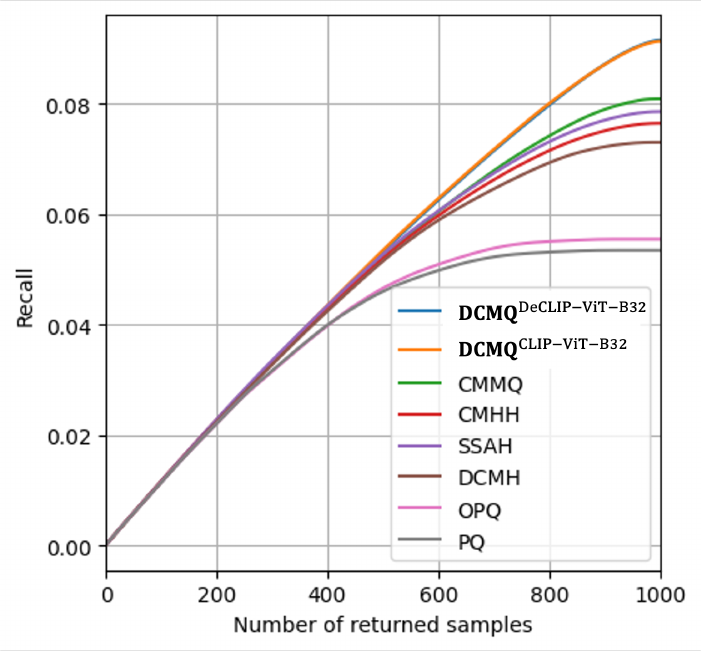}}
\caption{Top 1000-precision (a, b), and Recall at 1000 (c, d) curves on MIRFlickr @64-bit.}
\label{fig:Top1000precision,Recall100}
\vspace{-2em}
\end{figure*}

The experimental results on benchmarks are reported in Table \ref{table:Comparisons}, where the highest score for each setup is shown in bold, and we \colorbox{lime!15}{highlight} the performance of DCMQ. Notably, DCMQ produces the best mAP scores for all datasets in every bit-setup. Deep learning-based methods generally outperform non-deep ones (PQ, OPQ). For MIRFlickr, NUS-WIDE, and MS COCO, the average of mAP scores for all bit-setups yields differences of 30.3\%p, 44.4\%p, and 46.6\%p for the $I\rightarrow{T}$ task, and 21.4\%p, 43.6\%p, and 46.8\%p for the $T\rightarrow{I}$ task, between the non-deep methods and DCMQ respectively. Using the same backbone, RN18, when compared with deep methods (DCMH, SSAH, CMHH, CMMQ), DCMQ achieves 10.7\%p, 12.8\%p, and 20.1\%p for the $I\rightarrow{T}$ task, and 9.4\%p, 16.0\%p, and 31.3\%p for the $T\rightarrow{I}$ task higher retrieval scores on average. Note that DeCLIP uses more data transformation schemes than CLIP, while CLIP is trained with more image-text pairs. This seems to have the affect that the smaller gallery size of MIRFlickr favors $\text{DCMQ}^{\textrm{DeCLIP-ViT-B32}}$ while MS COCO with a larger gallery and more label categories favors $\text{DCMQ}^{\textrm{CLIP-ViT-B32}}$. In addition, referring to Figure \ref{fig:Top1000precision,Recall100}, DCMQ significantly outperforms others by large margins under two metrics, Top 1000-precision and recall at 1000. These results showcase DCMQ's potential for deployment in real cases where a large number of similar samples are retrieved with high precision and recall levels.


\subsection{Ablation Study}
\label{subsec:Ablation Study}

\begin{table}[!t]
\caption{DCMQ (RN18 and 3 MLP layers as student) with different teachers: CLIP \cite{CLIP}, DeCLIP \cite{DeCLIP}, ALBEF \cite{ALBEF} and BLIP \cite{BLIP} @64-bit.}
\centering
\begin{adjustbox}{width=0.99\textwidth}
\begin{tabular}{c|c|c|c|c||c|c|c|c|c}
\toprule
Task & Teacher & MS COCO & MIRFlickr & NUS-WIDE & Task & Teacher & MS COCO & MIRFlickr & NUS-WIDE \\ \midrule
\multirow{9}{*}{$T\rightarrow{I}$}
& CLIP-RN50 & 0.861 & 0.816 & 0.763 & \multirow{9}{*}{$I\rightarrow{T}$} & CLIP-RN50 & 0.837 & 0.844 & 0.758 \\ \cmidrule{2-5} \cmidrule{7-10}
& CLIP-ViT-B32 & 0.859 & 0.815 & 0.761 & & CLIP-ViT-B32 & 0.837 & 0.851 & 0.764 \\ \cmidrule{2-5} \cmidrule{7-10}
& CLIP-ViT-L14 & \textbf{0.878} & 0.812 & 0.770 & & CLIP-ViT-L14 & 0.854 & 0.846 & 0.747 \\ \cmidrule{2-5} \cmidrule{7-10}
& DeCLIP-RN50 & 0.849 & 0.814 & 0.759 & & DeCLIP-RN50 & 0.829 & 0.846 & 0.739 \\ \cmidrule{2-5} \cmidrule{7-10}
& DeCLIP-ViT-B32 & 0.856 & 0.818 & 0.761 & & DeCLIP-ViT-B32 & 0.826 & 0.855 & 0.747 \\ \cmidrule{2-5} \cmidrule{7-10}
& ALBEF & 0.866 & 0.816 & 0.768 & & ALBEF & 0.852 & 0.853 & 0.764 \\ \cmidrule{2-5} \cmidrule{7-10}
& BLIP w/ ViT-B & 0.872 & \textbf{0.819} & \textbf{0.772} & & BLIP w/ ViT-B & \textbf{0.856} & \textbf{0.856} & \textbf{0.766} \\ \bottomrule
\end{tabular}
\end{adjustbox}
\label{table:different inference protocols}
\vspace{-1em}
\end{table}

\begin{table}[!t]
\caption{Investigation on impact of NPC with CLIP-ViT-B32 as the teacher VLP @64-bit.}
\centering
\begin{adjustbox}{width=0.99\textwidth}
\begin{tabular}{c|c|c|c|c||c|c|c|c|c}
\toprule
\multirow{2}{*}{Task} & \multirow{2}{*}{Teacher} & \multirow{2}{*}{MS COCO} & \multirow{2}{*}{MIRFlickr} & \multirow{2}{*}{NUS-WIDE} & \multirow{2}{*}{Task} & \multirow{2}{*}{Teacher} & \multirow{2}{*}{MS COCO} & \multirow{2}{*}{MIRFlickr} & \multirow{2}{*}{NUS-WIDE} \\ & &  & & & & & & & \\ \midrule
\multirow{5}{*}{$T\rightarrow{I}$}
& Identity & 0.801 & 0.788  & 0.740 & \multirow{5}{*}{$I\rightarrow{T}$} & Identity & 0.792 & 0.818 & 0.758 \\ \cmidrule{2-5} \cmidrule{7-10}
& With multi-hot labels & 0.821 & 0.794 & 0.723 & & With multi-hot labels & 0.820 & 0.824 & 0.720 \\ \cmidrule{2-5} \cmidrule{7-10}
& Without NPC & 0.521 & 0.747 & 0.592 & & Without NPC & 0.628 & 0.791 & 0.606 \\ \cmidrule{2-5} \cmidrule{7-10}
& With NPC  & \textbf{0.859} & \textbf{0.815}  & \textbf{0.761} & & With NPC & \textbf{0.837} & \textbf{0.851}  & \textbf{0.764} \\ \bottomrule
\end{tabular}
\end{adjustbox}
\label{table:effect of NPC}
\vspace{-1em}
\end{table}

\paragraph{The impact of knowledge distillation with VLP teachers.}
We conduct ablations on using different VLP teachers for DCMQ and list the results in Table \ref{table:different inference protocols}. For CLIP \cite{CLIP} and its variants, we employ different image encoder backbones namely RN50: ResNet50, ViT-B32, and ViT-L14. Similarly, for DeCLIP \cite{DeCLIP}, we choose the same RN50 and ViT-B32 backbone models as teacher. We also demonstrate that DCMQ offers the flexibility of accommodating structurally different VLP models, including those using inference protocols different from CLIP (ALBEF \cite{ALBEF} and BLIP \cite{BLIP}). As long as image and text embeddings can be computed to construct the similarity matrix, DCMQ is applicable. In fact, using BLIP with ViT-B produces even stronger results than using CLIP. This demonstrates that utilizing a higher-performing VLP model as a teacher generally increases retrieval accuracy, thus validating the usefulness of distilling knowledge from VLP.

\begin{table}[!t]
\begin{minipage}{0.47\textwidth}
\caption{Impact of Gumbel-Softmax (in Equation \ref{equation:GumbelSoftmax}) on MIRFlickr.}
\centering
\begin{adjustbox}{width=\textwidth}
\begin{tabular}{c|c|c|c|c}
\toprule 
Task & Type & 32-bit & 64-bit & 128-bit \\ \midrule
\multirow{2}{*}{$T\rightarrow{I}$} & with Gumbel & 0.814 & 0.816 & 0.811 \\ \cmidrule{2-5}
 & without Gumbel & 0.805 & 0.804 & 0.795 \\ \midrule
 \multirow{2}{*}{$I\rightarrow{T}$} & with Gumbel & 0.845 & 0.844 & 0.845 \\ \cmidrule{2-5}
 & without Gumbel & 0.832 & 0.835 & 0.836 \\
\bottomrule
\end{tabular}
\end{adjustbox}
\label{table:impact_gumbel}
\end{minipage}\hfill
\begin{minipage}{0.48\textwidth}
\caption{Efficiency evaluation on MS COCO.}
\centering
\begin{adjustbox}{width=\textwidth}
\begin{tabular}{c|c|c|c}
\toprule 
 & CLIP & CMMQ & DCMQ \\ \midrule
Storage per sample (byte) & 1,024 & 8 & 8 \\ \midrule 
Storage for gallery (mb) & 125.05 & 0.98 & 0.99 \\ \midrule 
Model parameters (M) & 102.27 & 26.55 & 26.73 \\ \midrule 
Training time (sec)  & 2231.22 & 359.10 & 364.58 \\ \midrule 
mAP: $T\rightarrow{I}$ & 0.854 & 0.559  & 0.865 \\ \midrule
mAP: $I\rightarrow{T}$ & 0.834 & 0.648  & 0.842 \\
\bottomrule
\end{tabular}
\end{adjustbox}
\label{table:practicality evaluation}
\end{minipage}
\end{table}





\paragraph{The impact of NPC.} Table \ref{table:effect of NPC} shows the influence of the NPC on the performance. Here, `Identity' refers to the use of an Identity matrix of size $N$ as a substitute for $\mathbf{T}$. The term `With multi-hot labels' denotes the construction of a target similarity matrix using multi-hot labels. `Without NPC' indicates the exclusion of Algorithm \ref{algorithm:NPC} during the training phase and the application of pure similarity scores instead. The results clearly demonstrate that the combination of VLP teachers with NPC outperforms all other setups. Interestingly, direct application of VLP knowledge without NPC tends to degrade performance. However, incorporating NPC significantly enhances performance, highlighting the essential role of NPC in the distillation process.

\paragraph{The impact of PQG.}  In addition, as shown in Table \ref{table:impact_gumbel}, using PQG to regularize codewords is an effective measure that can be utilized to boost performance further. This demonstrates the advantages of PQ in representing similarity scores as real values as opposed to hashing. The entire DCMQ proposals show the best results when combined all together, verifying their effectiveness.

\paragraph{Cost and efficiency comparison.} We run experiments to demonstrate the superiority of DCMQ in terms of computational cost and performance, comparing the computational costs and mAP achieved using a 128-D floating type fine-tuned CLIP (CLIP-RN50), a 128-bit binary CMMQ (RN18), and a 128-bit DCMQ (RN18 as student CLIP-RN50 as teacher). In Table~\ref{table:practicality evaluation}, we can see that both CMMQ and DCMQ, which approximate nearest neighbors in Hamming space to increase retrieval speed, also significantly reduces storage costs over 120 times, compared to floating type gallery embeddings in CLIP. In terms of mAP, both CLIP and DCMQ outperform CMMQ which does not utilize VLP, but most importantly, DCMQ produces the best performance in spite of the quantization. Additional results directly comparing trained CLIP with DCMQ can be found in Appendix \ref{sec:appendix:Direct utilization of a small size VLP}.

As DCMQ operates with approximate nearest neighbor search, as outlined in Section \ref{sec:Inference}, it not only has lower storage costs but also attains a faster search speed. These advantages become substantially more pronounced as the gallery size increases. Furthermore, our method needs only a single inference of VLP during data preparation and utilizes the same backbone architecture as previous learning to hash methods, thereby incurring minimal additional training costs with high return.

\begin{figure}[!t]
\centering
\includegraphics[width=0.9\linewidth]{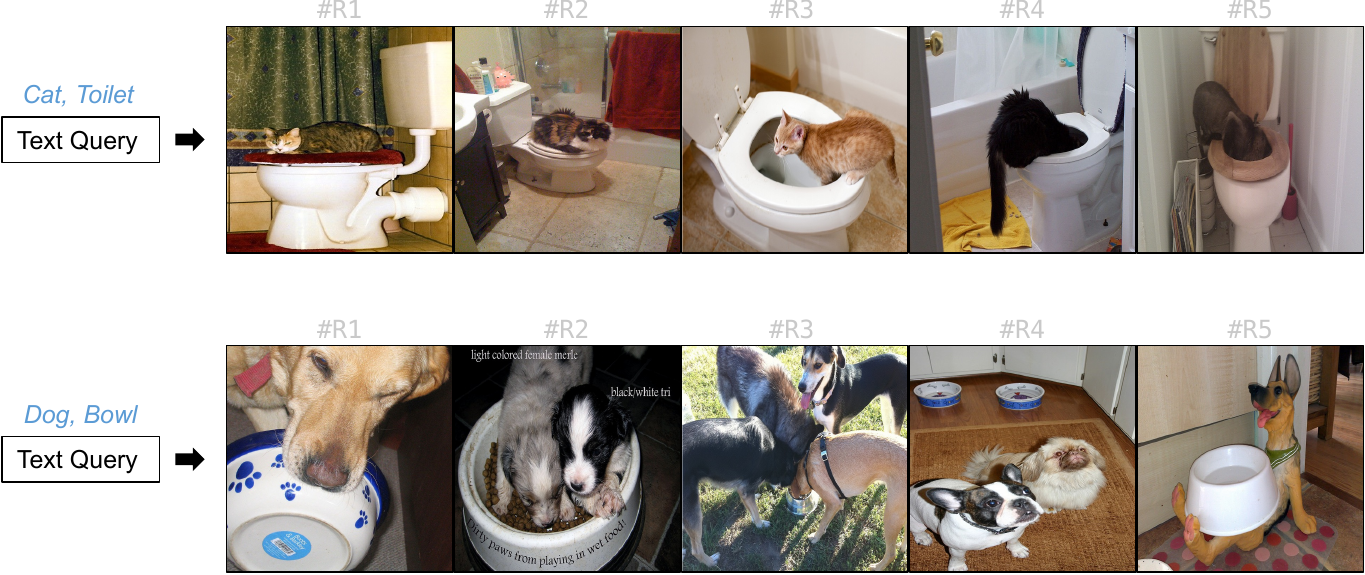}
\caption{Retrieval results with DCMQ hash codes on MS COCO @128-bit. Text query is given to retrieve images in the gallery. Label names are written in blue.}
\label{fig:retrieval_results}
\end{figure}

\paragraph{Qualitative Results.} We examine the qualitative results of DCMQ, which has been trained using CLIP-ViT-L14 as the teacher. The results, depicted in Figure \ref{fig:retrieval_results}, clearly demonstrate that DCMQ produces high-quality retrieval outcomes.

\section{Discussion \& Conclusion}

\paragraph{Limitation} While DCMQ shows promise in retrieval tasks, its effectiveness in other types of tasks remains untested. These limitations point towards future research opportunities in expanding DCMQ's applicability. Moreover, regarding potential social impacts and dangers, retrieval systems may raise privacy concerns due to their reliance on training data, including sensitive information, leading to potential discriminatory outcomes. Therefore, we need to take careful measures to mitigate these.

\paragraph{Conclusion} This paper presents Distillation for Cross-Modal Quantization (DCMQ), a novel method that leverages the semantic knowledge of VLP models to enhance hashing-based cross-modal retrieval. We introduce two innovative techniques: Normalization with Paired Consistency (NPC) for effective utilization of learned image-text similarity, and Product Quantization with Gumbel (PQG) for robust quantization. DCMQ achieves state-of-the-art performance on three renowned benchmarks, as confirmed by extensive experiments. This work represents the first successful attempt to apply VLP-inherent semantics to quantization-based cross-modal retrieval.

{\small
\bibliographystyle{ieee_fullname}
\bibliography{custom}
}

\newpage

\appendixpage

\section{Dataset Configuration}
\label{sec:appendix:dataset}

\textit{MS COCO} \cite{MSCOCO} is an image-text paired dataset consisting of over 130K samples where each sample is assigned one or more semantic labels out of 80 categories. As with other methods \cite{SCAHN,CMMQ}, we randomly respectively selected 10,000 and 5,000 samples for training and test, and utilize 2026-dimensional BoW as text data. Since MS COCO set has actual text descriptions (captions) per image, we also evaluate DCMQ with real text inputs by replacing text encoder with VLP's text Transformer \cite{CLIP}, in Sub-Section \ref{sec:appendix:Beyond Bag of Words}.

\textit{NUS-WIDE} \cite{NUS-WIDE} contains over 260K images gathered from the Web, all annotated with at least one conceptual label out of 81. Just like previous works \cite{DCMH,CMHH,CMMQ}, we filter out images without text description and choose images belonging to the 21 most frequent concepts for evaluation. The training set and the query set comprise 500 and 100 images respectively for every concept selected, each of which is associated with a 1,000-dimensional text BoW.

\textit{MIRFlickr} \cite{MIRFlickr} consists of 25K images collected from Flickr, each of which has multiple class labels. These labels are utilized by the VLP teacher as well as for evaluation. Each image is also associated with multiple common captions out of 1,386, but they are not specifically provided for each image. Instead, a 1,386-dimensional Bag of Words (BoW) feature vector is given per image. In total, following the dataset preparation procedure in \cite{DCMH}, we end up with 20,015 samples, where the text is represented by the BoW feature vector.

\section{Extremely short bit}
\label{sec:appendix:Under more difficult condition}

To demonstrate the efficacy of DCMQ, we perform further experiments under the challenging conditions of building a retrieval system with an exceptionally low number of bits. The number of codebooks is set to $M={2,4}$, with results presented in Table \ref{table:Extremely Short bit}. Employing a PQ-based learning to hash method enables us to incorporate robust representations like PQG and NPC during training, thereby achieving comparable performance even under minimal bit conditions.

\begin{table}[!h]
\centering
\caption{mAP(@5000) scores for DCMQ (RN18 and 3MLP layers as student) with extremely short bits. Despite a significant reduction in bits, there is only a minimal decrease in accuracy.}
\begin{adjustbox}{width=0.6\textwidth}
\begin{tabular}{c|c|c|c|c|c|c}
\toprule

\multirow{2}{*}{Teacher}      & \multicolumn{2}{c|}{MS COCO} & \multicolumn{2}{c|}{MIRFlickr} & \multicolumn{2}{c}{NUS-WIDE} \\ \cmidrule{2-7} 
             & 8-bit       & 16-bit       & 8-bit        & 16-bit       & 8-bit      & 16-bit      \\ \midrule 

\multicolumn{7}{c}{Task: $T\rightarrow{I}$}  \\ \midrule             
CLIP-RN50    & \textbf{0.772}                      & \textbf{0.833}   & 0.789                     & 0.811                      & 0.705                     & 0.731                                        \\
CLIP-ViT  & 0.769                     & 0.820    & 0.783                     & 0.804                      & 0.704                     & \textbf{0.737}                                            \\
DeCLIP-RN50  & 0.754                     & 0.822     & 0.796                     & 0.812                      & \textbf{0.711}                      & 0.734                                       \\

DeCLIP-ViT & 0.752                     & 0.804   & \textbf{0.803}                     & \textbf{0.813}                      & 0.710                       & 0.732                              \\ 

\midrule \midrule

\multicolumn{7}{c}{Task: $I\rightarrow{T}$}  \\ \midrule             
CLIP-RN50  & \textbf{0.767}     & \textbf{0.820}    & 0.826       & 0.835        & \textbf{0.747}        & \textbf{0.752}             \\
CLIP-ViT-B32    & 0.766      & 0.809    & 0.824       & 0.841        & 0.740         & 0.748           \\
DeCLIP-RN50     & 0.744      & 0.808  & 0.823       & 0.845        & 0.734        & 0.732          \\

DeCLIP-ViT-B32   & 0.745      & 0.789  & \textbf{0.835}       & \textbf{0.852}        & 0.725        & 0.744       \\

\bottomrule
\end{tabular}
\end{adjustbox}
\label{table:Extremely Short bit}
\end{table}

\clearpage

\section{Text description as input data}
\label{sec:appendix:Beyond Bag of Words}

\begin{figure}[!h]
\centering
  \subcaptionbox{$T\rightarrow{I}$
  \label{fig:gumbel_t2i}}{\includegraphics[height=3.5cm]{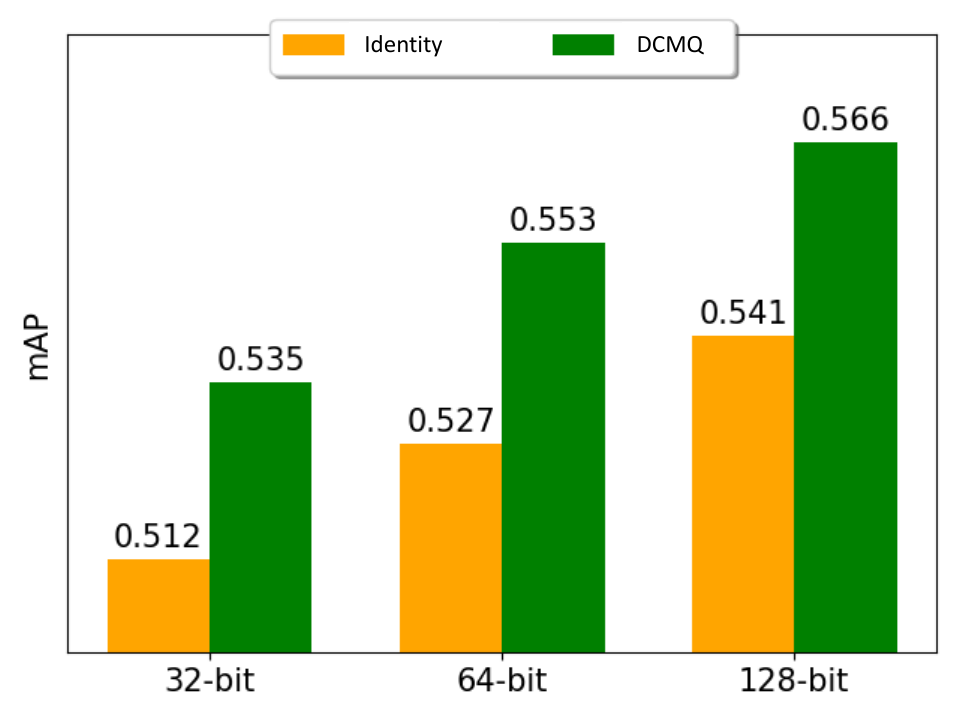}}
  \subcaptionbox{$I\rightarrow{T}$
  \label{fig:gumbel_i2t}}{\includegraphics[height=3.5cm]{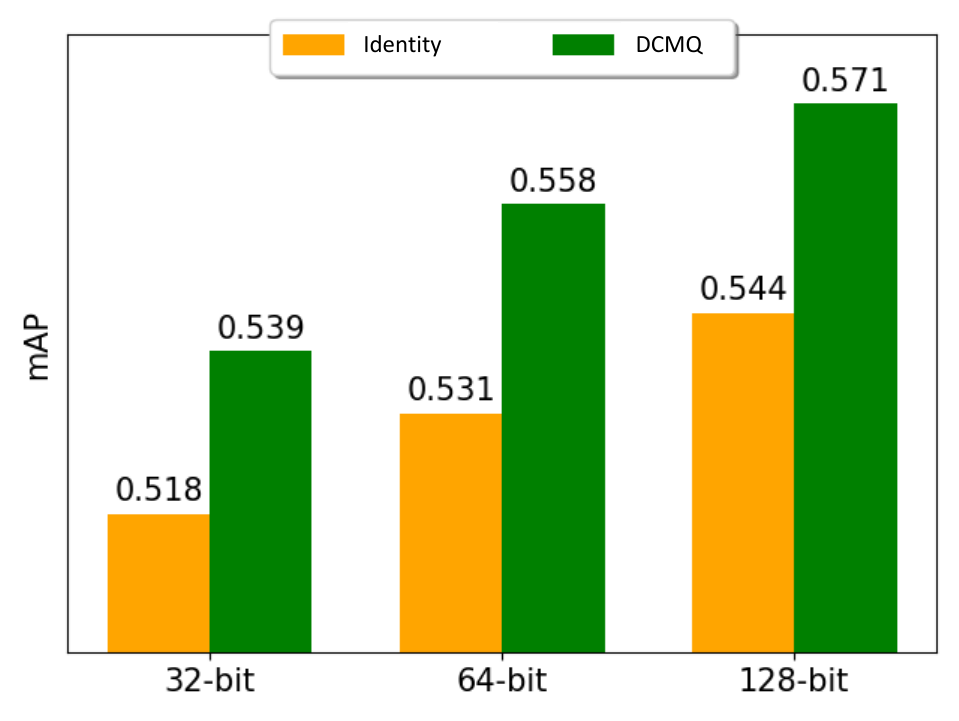}}
\caption{mAP(@5000) scores of DCMQ integrated with CLIP-RN50 text encoder, given text captions of MS COCO. For the teacher VLP, we use CLIP-ViT-B32.}
\label{fig:Hist caption}
\end{figure}

\begin{figure}[!h]
\centering
\includegraphics[width=0.8\linewidth]{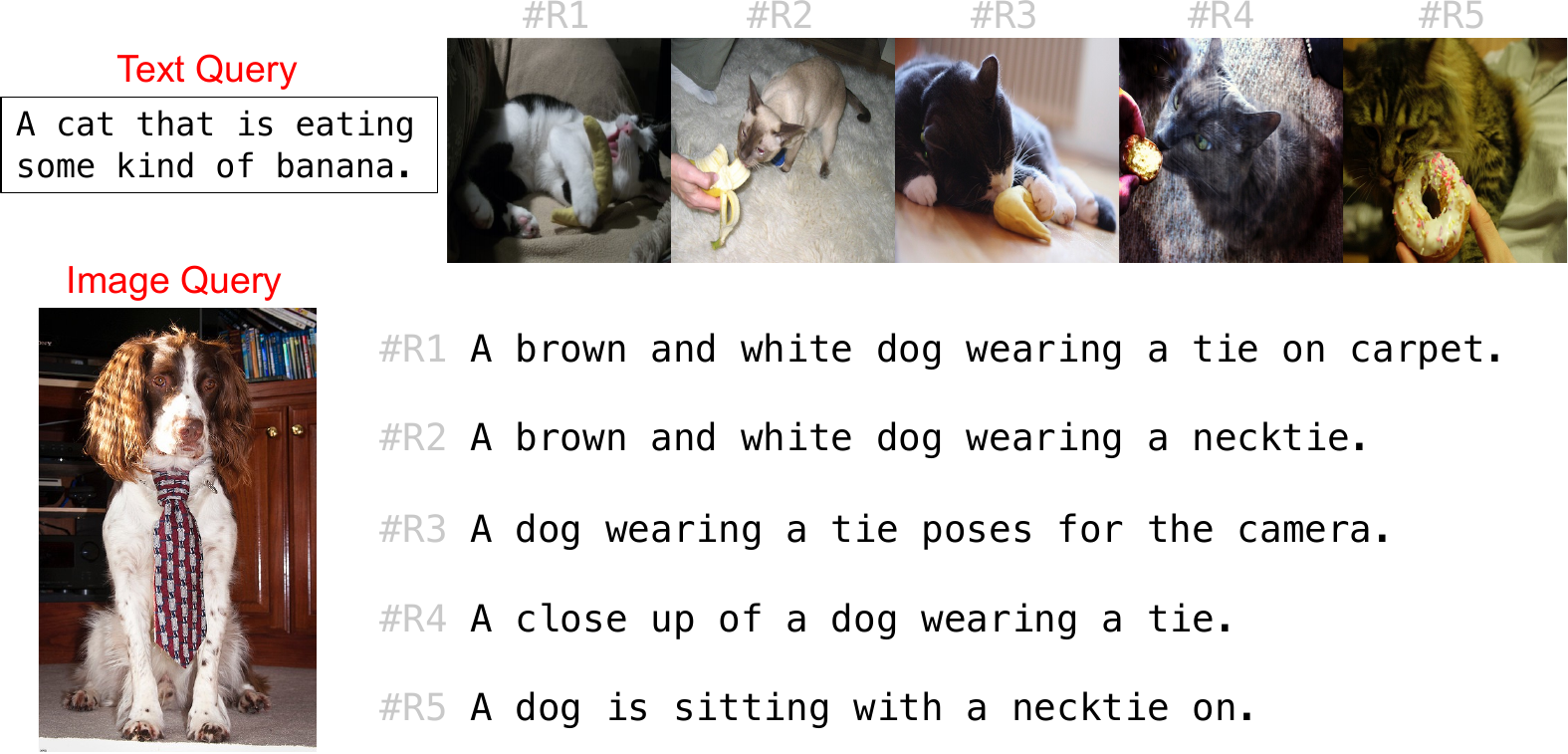}
\caption{Cross-modal retrieval results with DCMQ hash codes on MS COCO, using actual captions. }
\label{fig:Qualitative}
\end{figure}

In this work, we opt for MLPs as the text encoder for the student model, a choice driven by the benchmarks' reliance on Bag-of-Words (BoW) feature vectors rather than raw text. However, a question arises as: \textit{What if the actual text is utilized?}. Therefore, we employ a text encoder in the student model that can process actual text, to explore its impact on hashing performance. Given that the MS COCO dataset includes detailed text descriptions (captions) for each image, we utilize the Vision-Language Pretraining (VLP) of CLIP-RN50 \cite{CLIP} for the student model, inputting captions into the text encoder. Starting with both the VLP's image and text encoders, we assess the effectiveness of the DCMQ approach, which uses captions instead of BoW vectors for retrieval. The results, as illustrated in Figure \ref{fig:Hist caption}, demonstrate that compared to the Identity (self-supervised) setting, the DCMQ training scheme also enhances VLPs in the cross-modal retrieval of images and actual text. Furthermore, we examine the qualitative results of DCMQ trained with actual text as depicted in Figure \ref{fig:Qualitative}. It is evident that DCMQ generates high-quality retrieval results.

\clearpage

\section{Different image encoder backbones}
\label{sec:appendix:Different image encoder backbones}

We explore the potential of DCMQ with stronger image backbones for the student in Table \ref{table:Backbone change}, which can be useful when practitioner may be willing to increase computational load for the sake of stronger performance. Specifically, we investigate the effects of using the following as the image encoder: pretrained AlexNet \cite{AlexNet}, and ResNet50 (RN50) \cite{ResNet} with its variants: DINO-RN50 \cite{DINO}, CLIP-RN50 \cite{CLIP}, and Vision Transformer (ViT-B32) \cite{ViT} with its variant: DeiT \cite{DeiT}. We fix the teacher of DCMQ as CLIP-RN50 to ensure a fair comparison.

Not surprisingly, as we can observe from the results, the stronger the backbone (DeiT is the strongest one when considering the classification task), the better the performance in general. Note that even though we employ randomly initialized simple 3-layer text encoder here, all cross-modal retrieval performances improve along with more powerful image encoder backbones. For CLIP-RN50 backbone, the scores are a bit lower than expected, and we suspect that is because the teacher is also using CLIP-RN50 as mentioned and the additional knowledge transfer may not be as significant. On the other hand, interestingly, even for AlexNet, DCMQ achieves comparable results.

\begin{table*}[!h]
\centering
\caption{mAP(@5000) scores of DCMQ with different backbone architectures as image encoder.}
\begin{adjustbox}{width=0.99\textwidth}
\begin{tabular}{c|l|c|c|c|c|c|c|c|c|c}
\toprule
\multirow{2}{*}{Task}    &  \multicolumn{1}{c|}{\multirow{2}{*}{Backbone}} & \multicolumn{3}{c|}{MS COCO} & \multicolumn{3}{c|}{MIRFlickr} & \multicolumn{3}{c}{NUS-WIDE} \\ \cmidrule{3-11} 
                           &    \multicolumn{1}{c|}{}                        & 32-bit   & 64-bit  & 128-bit  & 32-bit   & 64-bit  & 128-bit  & 32-bit  & 64-bit  & 128-bit  \\ \midrule
\multirow{6}{*}{$T\rightarrow{I}$}   & AlexNet \cite{AlexNet} & 	0.813	   & 0.824		  & 0.828  & 0.801	& 	0.804		  & 	0.802	 & 	0.739	    & 0.753   & 		0.761		     \\
                       &             RN50 \cite{ResNet}    & 0.888  & 	0.895	  & 	0.897     & 0.808	   & 0.812	   & 0.810		   & 0.753		  & 0.767	  & 	0.763		\\
                       &                  DINO-RN50 \cite{DINO}   & 0.871	   & 0.882  & 	0.888     & \textbf{0.817}	  & 	\textbf{0.817}		 & 0.818		  & 0.756		   & 0.765 & 	0.766	    \\
  &                  CLIP-RN50 \cite{CLIP} & 0.806		   & 0.828	  & 0.852 & 0.811	   & 0.813  & 		\textbf{0.820}	  & 0.748 &	0.750	    & 0.752	    \\
  &                  ViT-B32 \cite{ViT}     & 	0.897	  & 0.906		  & 0.908    & 0.813	   & 	0.815		   & 0.816	   & 0.761	   & 	0.766		   &0.765	  \\ 
&                 DeiT \cite{DeiT}    & 	\textbf{0.907}		  & \textbf{0.913}		   & \textbf{0.915}    & 0.813	    & 	0.814  & 		0.813		  & \textbf{0.768}		   & \textbf{0.769}  & \textbf{0.773}	   \\ 
\midrule 
\multirow{6}{*}{$I\rightarrow{T}$}   & AlexNet \cite{AlexNet}   & 	0.784		   & 0.792	  & 0.794 & 0.827	& 	0.828		  & 	0.830	  & 	0.731		    & 0.746	   & 	0.742		  \\
                       &             RN50 \cite{ResNet}   & 0.828	  & 0.879	  & 0.881     & 0.852	  & 0.855	   & 0.851		   & 0.762		  & 0.770	  & 0.778		  \\
                       &                  DINO-RN50 \cite{DINO}     & 0.862	   & 0.870		  & 0.873    & \textbf{0.862}   & 		0.859	 & 	0.859		  & 0.769		   & 0.784	   & 	0.783	     \\
  &                  CLIP-RN50 \cite{CLIP}  & 	0.780  & 	0.781 & 	0.795 & 0.835	   & 	0.830	 & 	0.809	  & 	0.718	& 	0.745	    & 0.711   \\
  &                  ViT-B32 \cite{ViT}   & 		0.887		  & 0.892		  & 0.893    & 0.859	   & 	0.859	   & 	0.860	   & 	0.779		   & 0.785		   & 0.792  \\ 
&                 DeiT \cite{DeiT}  & \textbf{0.901}		  & \textbf{0.904}		   & \textbf{0.905}	     & 0.860		    & \textbf{0.864}	   & 	\textbf{0.861}		  & \textbf{0.785}	  &  \textbf{0.792}		  & \textbf{0.797}	      \\ 
\bottomrule
\end{tabular}
\end{adjustbox}
\label{table:Backbone change}
\end{table*}

\section{Effect of joint training}
\label{sec:appendix:Effect of joint training}

\begin{table}[!h]
\caption{Experimental results on joint training in Figure \ref{fig:Training}, with CLIP-ViT-B32 as Teacher @64-bit.}
\centering
\begin{adjustbox}{width=0.55\textwidth}
\begin{tabular}{c|c|c|c}
\toprule

Method     & MS COCO & MIRFlickr & NUS-WIDE \\ \midrule
\multicolumn{4}{c}{Task: $T\rightarrow{I}$}  \\ \midrule             
With joint training   & \textbf{0.859}                    & \textbf{0.815}               & \textbf{0.761}    \\
Without joint training   & 0.850                   & 0.809              & 0.755    \\

\midrule \midrule

\multicolumn{4}{c}{Task: $I\rightarrow{T}$}  \\ \midrule             
With joint training    & \textbf{0.837}                   & \textbf{0.851}               & \textbf{0.764}    \\
Without joint training    & 0.830                    & 0.843              & 0.758    \\ 

\bottomrule
\end{tabular}
\end{adjustbox}
\label{table:effect of joint training}
\end{table}

We investigate the effect of joint training in Table \ref{table:effect of joint training}. Similar to \cite{SPQ}, the joint training of the real-valued feature vector and the quantized vector enhances retrieval performance.

\clearpage

\section{Direct utilization of a small size VLP}
\label{sec:appendix:Direct utilization of a small size VLP}

\begin{table}[!h]
\caption{Experimental results on directly training a small size VLP @64-bit. We choose the CLIP-RN50 VLP model (which is the smallest model in CLIP family) for comparison with our RN18-based DCMQ.}
\centering
\begin{adjustbox}{width=0.45\textwidth}
\begin{tabular}{c|c|c|c}
\toprule

Method     & MS COCO & MIRFlickr & NUS-WIDE \\ \midrule
\multicolumn{4}{c}{Task: $T\rightarrow{I}$}  \\ \midrule             
Ours-RN18  & \textbf{0.861}                    & \textbf{0.816}               & \textbf{0.763}    \\
CLIP-RN50   & 0.820                 & 0.805              & 0.746    \\

\midrule \midrule

\multicolumn{4}{c}{Task: $I\rightarrow{T}$}  \\ \midrule             
Ours-RN18    & \textbf{0.837}                   & \textbf{0.844}               & \textbf{0.758}    \\
CLIP-RN50    & 0.779                  & 0.828              & 0.740    \\ 

\bottomrule
\end{tabular}
\end{adjustbox}
\label{table:direct VLP}
\end{table}

Directly fine-tuning smaller VLP models with multi-hot labels could show decent performance in learning to hash, similar to our report of CLIP in Table \ref{table:practicality evaluation}. However, these models inherently have limitations, which we further investigated by conducting additional experiments on CLIP-RN50. We fine-tuned it without distillation and applied standard supervised PQ \cite{PQN} to train 64-bit codebooks. The results, presented in Table \ref{table:direct VLP}, highlight these limitations.

Like most foundational models, the performance of VLP models scales proportionally with the size of the model and dataset, a phenomenon known as the scaling law. As a result, smaller VLP models are likely to underperform compared to their larger counterparts. Distillation offers a solution to this problem, allowing a smaller student model to leverage the power of a larger VLP. Our paper demonstrates that using high-performing VLP models as teachers can enhance retrieval performance (See Table \ref{table:different inference protocols}), even when the student model has a significantly smaller backbone, such as RN18.

\section{Impact of PQG on codeword distribution}
\label{sec:appendix:Histogram}

\begin{figure}[!h]
\centering
  \subcaptionbox{without PQG
  \label{fig:without_pqg}}{\includegraphics[height=3.3cm]{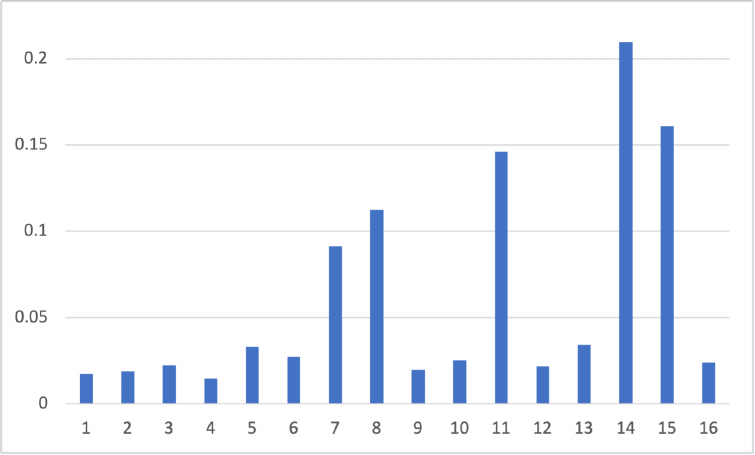}}
  \subcaptionbox{with PQG
  \label{fig:with_pqg}}{\includegraphics[height=3.3cm]{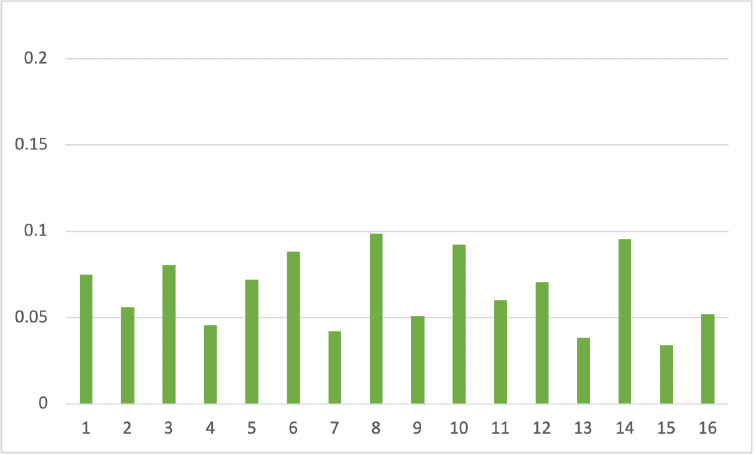}}
\caption{A histogram represents the indices of the selected codewords, with or without PQG, for the MIRFlickr @64-bit.}
\label{fig:PQG_existence}
\end{figure}

As shown in Fig. \ref{fig:PQG_existence}, without PQG training and when using the standard softmax operation, we notice that certain codewords tend to dominate. That is, most data samples map to these specific codewords, leading to low entropy and, hence, limited information capacity. Conversely, when PQG is applied, the codewords are selected more evenly, signifying an increase in entropy. This implies that the system can accommodate more information by selecting a variety of codewords, which, in turn, results in improved performance.

\end{document}